\documentclass{article}

\usepackage[preprint]{neurips_2023}
\usepackage{times}

\usepackage{amsmath,amsfonts,bm}

\def\eqref#1{equation~\ref{#1}}
\def\1{\bm{1}}

\DeclareMathAlphabet{\mathsfit}{\encodingdefault}{\sfdefault}{m}{sl}
\SetMathAlphabet{\mathsfit}{bold}{\encodingdefault}{\sfdefault}{bx}{n}

\usepackage[utf8]{inputenc} % allow utf-8 input
\usepackage[T1]{fontenc}    % use 8-bit T1 fonts
\usepackage{url}            % simple URL typesetting
\usepackage{booktabs}       % professional-quality tables
\usepackage{amsfonts}       % blackboard math symbols
\usepackage{nicefrac}       % compact symbols for 1/2, etc.
\usepackage{microtype}      % microtypography
\usepackage{xcolor}         % colors
\usepackage{multirow}
\usepackage{graphicx}
\usepackage{algorithm}
\usepackage{algorithmic}
\usepackage{caption}
\usepackage{subcaption}
\usepackage{hyperref}       % hyperlinks

\newcommand{\alg}{Fed-LPFM}

\title{Leveraging Foundation Models to Improve Lightweight Clients in Federated Learning}

\author{
  Xidong Wu\thanks{work done at BCAI} \\
  Electrical \& Computer Engineering \\
  University of Pittsburgh \\
  Pittsburgh, PA 15260 \\
  \And
  Wan-Yi Lin, Devin Wilmott, Filipe Condessa, Zhenzhen Li \& Madan Ravi Ganesh\thanks{corresponding author} \\
  BCAI \\
  2555 Smallman Street \\
  Pittsburgh, PA 15222 
  \And
  Yufei Huang \\
  Hematology/Oncology \\
  University of Pittsburgh\\
  Pittsburgh, PA 15260 \\
}
\begin{document}

\maketitle
\begin{abstract}
Federated Learning (FL) is a distributed training paradigm that enables clients scattered across the world to cooperatively learn a global model without divulging confidential data. 
However, FL faces a significant challenge in the form of heterogeneous data distributions among clients, which leads to a reduction in performance and robustness. 
A recent approach to mitigating the impact of heterogeneous data distributions is through the use of foundation models, which offer better performance at the cost of larger computational overheads and slower inference speeds. 
We introduce \textit{foundation model distillation} to assist in the federated training of lightweight client models and increase their performance under heterogeneous data settings while keeping inference costs low.
Our results show improvement in the global model performance on a balanced testing set, which contains rarely observed samples, even under extreme non-IID client data distributions.
We conduct a thorough evaluation of our framework with different foundation model backbones on CIFAR10, with varying degrees of heterogeneous data distributions ranging from class-specific data partitions across clients to dirichlet data sampling, parameterized by values between 0.01 and 1.0. 
\end{abstract}

\section{Introduction}
\label{sec:introduction}
% FL: what is FL
Federated learning (FL) is a decentralized training paradigm in machine learning~\citep{mcmahan2017communication} that trains one global model across multiple clients while preserving the privacy of client data.
A typical FL framework consists of a central server coordinating global model training by periodically aggregating clients' local models that are trained with locally-stored data.
% Core problems in FL: heterogeous data + communication / inference budget
Similar to a variety of distributed learning approaches, one of the major challenges for FL is that locally-stored client data are heterogeneous, which can result from uneven distributions or unbalanced patterns~\citep{li2020federated1}. 
This results in client-drift, commonly-seen accuracy drops, and non-convergence~\citep{zhao2018federated, karimireddy2020scaffold, hsieh2020non, li2020federated1}. 
In addition, many FL clients in real-world deployments are edge devices which have strict limitations on inference speeds and compute. 
This, in turn, restricts clients to using small-scale models for inference. 

% Foundation Models: A Potential Solution. WHY
To maintain model performance under heterogeneous data distributions, foundation models~\citep{bommasani2021opportunities} present a potential solution. 
Their widely known benefits, such as their comprehensive knowledge, transferable representations across a broad range of downstream tasks~\citep{radford2021learning,wang2022image}, and strong robustness to distribution shifts~\citep{ma2021simple} make them a strong candidate to mitigate the effects of heterogeneous distributions.
With this in mind, multiple recent methods explore fine-tuning foundation models under federated settings~\citep{qu2022rethinking, chen2022importance, guo2023promptfl, chen2022fedtune, su2022cross, guo2023pfedprompt}.
% Roadblocks 
Among these, \cite{chen2022fedtune} have shown that under extreme non-IID conditions, federated fine-tuning of foundation models forces a performance worse than training on local data only. 
In addition, these methods incur a relatively large increase in inference time and compute by directly using foundation models instead of small-scale alternatives like EfficientNet~\cite{tan2019efficientnet} or MobileNet~\cite{sandler2018mobilenetv2} for inference.

% What we want to address
In order to effectively leverage the performance of foundation models under non-IID data distributions while using small-scale backbones for faster inference, we propose an approach to \textit{distill} knowledge from foundation models into the small-scale client models (proxy models).
During training, foundation models are not directly fine-tuned, but rather are leveraged to update each client's proxy model; then proxy model updates are shared with and aggregated by the server model. At inference, only the proxy models are used.
Thus, the proxy models offer low latency inference while knowledge from the foundation models helps reduce the bias and diversify the knowledge of the proxy models, especially under heterogeneous local data distributions.
In addition, our proposed method is agnostic to the number and size of foundation models available to each client. This offers the option of personalization for each client, which can select the appropriate foundation model(s) based on the amount of storage and compute available, as well as local data characteristics; this is particularly beneficial when some clients have much more/less data than others.
We use the concept of personalization to highlight important directions for future work.

Overall, our main contributions are as follows: 
\begin{itemize}
    % FMFL + small backbone training (distillation)
    \item This is the first approach to leverage foundation models in FL via distillation to help improve the performance and robustness achievable in small-scale client models (\emph{e.g.} relative increase of $9.22\%$ for EfficientNetB0, $8.69\%$ for ResNet18 and $24.60\%$ for MobileNetV2).

    % Non-iid setting
    \item Within the space of low latency models, we provide a federated learning solution robust to various heterogeneous client data. Our approach outperforms prior art across a variety of client data distributions, from IID to various parametrized dirichlet distributions and class-specific partitions. 

    \item We explore the impact of leveraging representations from fine-tuned foundation models on local data versus pre-trained foundation models. Our results show that under IID data distributions, an initial step of fine-tuning foundation models offers no benefit over 0-shot foundation models, and significantly hinders accuracy as data heterogeneity increases, suggesting that directly fine-tuning foundation models leads to biased representations.
    
    % Personalization
    \item Our framework is also the first to allow clients the flexibility in choosing their locally-stored foundation models (personalization) according to the scale of compute and data available. We study the impact of variable foundation model backbones and highlight the importance of combining disparate feature representations correctly.
\end{itemize}
\section{Related Work}
\label{sec:related_work}
\paragraph{Federated Learning with Heterogenous Client Data} 
Federated learning is a distributed machine learning scheme which enables multiple clients to train a shared model while keeping their data private. 
Typically a central server federates the training procedure by periodically aggregating model updates from clients~\citep{mcmahan2017communication}.
% local, thereby preserving data privacy. 
Frequently, client data can have non-identical distributions which causes naive aggregation methods to not be able to guarantee global model convergence to a local minimum~\citep{zhao2018federated, li2020federated, hsieh2020non,li2020federated1}. 
To tackle this challenge, FL-algorithms such as FedProx~\citep{li2020federated1} add a proximal term to the local training objective to protect models in each client from over-fitting to the local data distribution; other approaches such as regularization~\citep{t2020personalized}, model mixture~\citep{deng2020adaptive, mansour2020three, hanzely2020federated}, clustering clients~\citep{sattler2020clustered, cho2021personalized}, multi-task learning~\citep{smith2017federated}, and meta-learning~\citep{fallah2020personalized} have been introduced to stabilize the trained models. 
In this work, we tackle the issue of heterogeneous data distributions by distilling knowledge from foundation models to proxy models, to help mitigate this issue without the need for additional data.

\paragraph{Foundation Models in FL} 
The past few years have witnessed the rapid development of foundation models with the integration of language~\citep{radford2018improving, devlin2018bert,radford2021learning}, vision~\citep{bao2021beit}, and audio modalities~\citep{tang2023any} across many tasks. 
In FL, foundation models have been used to improve the robustness of clients to distribution shifts and heterogeneous data distributions~\citep{qu2022rethinking} or the overall performance of the system~\citep{chen2022fedtune, guo2023promptfl, zhao2023fedprompt, lu2023fedclip, guo2023pfedprompt}.
However, existing works do not fully address the increase in computational overhead nor inference time that follow the use of foundation models.
In addition, even compressed foundation models~\citep{sanh2019distilbert,wu2023tinyclip} do not fully match the latency requirements of clients, which hinders their deployment in real-world settings.
Therefore, we propose the use of small-scale proxy models and distillation to leverage the performance of foundation models while keeping inference costs low.

\paragraph{Distillation}
Knowledge distillation is a teaching technique that transfers valuable insights and generalization capabilities from a trained teacher model to a student model~\citep{hinton2015distilling,anil2018large, zhang2018deep, zhang2021adversarial}. 
Within the domain of FL, \cite{lin2020ensemble} explore adaptable aggregation methods with ensemble distillation at the server, while \cite{sattler2021fedaux} use an auxiliary dataset to weight and ensemble local models from each client.
FedDistill~\citep{seo202216} extracts statistics related to the logit-vector from different client models and shares them with the remaining clients to help with distillation. 
\cite{zhu2021data} present a data-free knowledge distillation approach by training a generative model at the server, using information from clients. 
They proceed to use the generative model to create synthetic data which is used to train client models. 
\cite{cho2021personalized} propose a co-distillation-based personalized FL method to allow cross-architecture training.
In our approach, we study the impact of knowledge distillation~\cite{hinton2015distilling} on the performance of small-scale client models without the use of excessive data, augmentations or model sharing so as to maintain privacy.
We hope to provide guidance with respect to how foundation models can be effectively used in FL. 
\section{Distilling Foundation Models in Federated Learning}

\subsection{Federate Learning: Setup}
Our core FL scheme follows FedAvg~\citep{mcmahan2017communication}, which consists of a central server and multiple clients, indexed as $i=1,2,...,N$. 
Each client-$i$ has its local private dataset $\mathcal{D}_i$. 
We denote the local loss function of interest for the $i$-th client as $\mathcal{L}(D_i; \theta)$, where $\theta \in \mathbb{R}^d$ are the parameters of the trainable client model. 
The overall optimization problem considered at the server is denoted as,
\begin{equation}
\label{eq:1}
\min _{\theta \in \mathbb{R}^{d}}\mathcal{L}(\theta):= \sum_{i=1}^{N}p_i \mathcal{L}(D_i; \theta).
\end{equation}
Here, $p_i$ is a re-weighting factor conditioned as  $p_i \geq 0$ and $\sum_{i} p_i = 1$. 
Typically, $p_i$ is assigned as $p_i = \frac{|\mathcal{D}_{i}|}{\sum_{j\in\mathcal{S}_t} |\mathcal{D}_{j}|}$ where $S_t$ denotes the set of clients communicating with the server at round $t$.
With this setup in mind, the FL framework repeats the following steps until a desired end condition is achieved:  1) The server broadcasts the current global model to selected clients; 2) Each client resets its local model with the received model, performs local training based on its data, and sends the updated weights/gradients to the server; 3) The central server updates the global model by aggregating the received weights/gradients. 

\begin{algorithm}[tb] 
\caption{\alg}
\label{alg:1}
\begin{algorithmic}[1]
\STATE {\bfseries Input:} Dataset $\mathcal{D}_i$, frozen and private pre-trained foundation models: $\mathcal{M}^1_i, \mathcal{M}^2_i, \ldots, \mathcal{M}^{M_i}_i$ and proxy model $\theta_0$ for each client $i \in [N]$.
\STATE
% \STATE {Initialize proxy models with $\theta_0$.} 
\STATE {\bfseries Server:} 
\FOR{Round $t = 0, 1, 2, \ldots, T-1$}
\STATE Send $\theta_t$ to connected clients $\mathcal{S}_t \subset [N]$. Let $P_t=\sum_{i\in\mathcal{S}_t}|\mathcal{D}_i|$.
% \STATE {\bfseries Clients execute:} 
% \STATE \textit{Clients-end local training:} 
\FOR{Client $i \in \mathcal{S}_t$ in parallel}
\STATE  $\theta^i_{t}$  $\leftarrow$ \textbf{LocalUpdate}($\theta_{t}$, $i$)
\STATE Send the updated model $\theta_t^i$ to the central server
\ENDFOR
\STATE Server-end aggregation: $\theta_{t+1} = \sum_{i\in\mathcal{S}_t} \frac{|D^i|}{P_t} \theta^i_{t}$
\ENDFOR
\STATE {\bfseries return: } $\theta_{T}$ 

\STATE
\STATE {\bfseries LocalUpdate}($\theta_{t}, i$)
\STATE $\theta^{i}_{t} = \theta_{t}$
% \FOR{Epoch $q = 0, 2, \ldots, Q-1$}
% \STATE $\theta^{i}_{t, q+1} = \theta^{i}_{t,q} - \eta \tilde{\nabla} \mathcal{L} (\theta^{i}_{t,q}; \mathcal{M}^i_{1},\ldots, \mathcal{M}^i_{M_i}; \mathcal{D}_i)$
% \ENDFOR
\STATE \textbf{for} epoch $q = 0, 1, \ldots, Q-1$: $\theta^{i}_{t, q+1} = \theta^{i}_{t,q} - \eta \tilde{\nabla} \mathcal{L} (\theta^{i}_{t,q}; \mathcal{M}^i_{1},\ldots, \mathcal{M}^i_{M_i}; \mathcal{D}_i)$
\STATE {\bfseries return: } $\theta^i_t = \theta^i_{t,Q}$ 

\end{algorithmic}
\end{algorithm}
\subsection{\alg{}}

\paragraph{Setup}
Unique to our framework, we consider the scenario where each client has access to local pre-trained foundation models. Similar to each client's training dataset, these foundation models are only accessible by the client and not other entities in FL.  
We assume that in the FL system each client contains two sets of local models: (a) a set $M_i$ of pre-trained foundation models (private): $\mathcal{M}^1_i, \mathcal{M}^2_i, \ldots, \mathcal{M}^{M_i}_i$, and (b) one trainable small-scale proxy model parameterized by $\theta_i$. 
Since the foundation models are private, only the proxy models are circulated among the clients and server to facilitate the exchange of knowledge across the system. 
Our goal is to minimize the objective in Eq.~\ref{eq:1}, where the $\theta$ to be optimized represents the parameters of the small-scale proxy model while the foundation models are left unmodified. 

\paragraph{Local Training}
% We specifically modify the loss function used during the local training stage within each client.
% Typically, during the client's local training stage in each round the client first receives an updated small-scale model from the server, which it then uses to reset its own proxy model and proceed with training on its local data. 
In our algorithm, the client uses its locally stored data along with the knowledge from its private foundation models to supervise local training. 
For this purpose we use the following loss function,
% More specifically, the client updates it's proxy model by minimizing the following loss function:
\begin{align} 
% \label{eq:2}
\mathcal{L}(D_i; \theta) &= \lambda \mathcal{L}_{CE} (D_i; \theta) + (1  - \lambda) \mathcal{L}_{Distill} (D_i; \theta, \mathcal{M}^1_i, \ldots, \mathcal{M}^{M_i}_i).
\end{align}
Here, the first term is the local cross entropy loss, denoted as
\begin{equation}
    \mathcal{L}_{CE}(D_i; \theta) =  \mathbb{E}_{(x,y) \sim \mathcal{D}_i} \ell_{CE} (h(x; \theta), y),
\end{equation}
where $h(\cdot)$ denotes the outcome of a forward pass through the proxy model.
The second term $\mathcal{L}_{Distill}$ is used to distill the knowledge between the proxy model and the pre-trained foundation models. 
Typically, the Kullback Leibler (KL) Divergence loss is used for this purpose. 
\begin{equation}
    \mathcal{L}_{Distill} (D_i; \theta, \mathcal{M}^1_i, \ldots, \mathcal{M}^{M_i}_i)= \sum_{m=1}^{M_i}\mathbb{E}_{(x,y) \sim \mathcal{D}_i}  \ell_{KL}[h(x; \theta) || \mathcal{M}^m_i (x)]
\end{equation}
The parameter $\lambda$ controls the proportion of knowledge distilled from foundation model in comparison to ground-truth labels.

\textbf{Aggregation scheme}
After local training, the server synchronizes with the available clients and aggregates the locally updated proxy models. 
The local models are aggregated with the following re-weighting scheme,
\begin{equation}
   \theta_{t+1} = \sum_{i \in S_t} \frac{|D_i|}{\sum_{j \in S_t} |\mathcal{D}_{j}|} \theta^i_t,
\end{equation} 
where $t$ denotes the communication round.
After the aggregation is complete, the server broadcasts the updated model to clients and the entire process is repeated until a desired end condition is met. 
% Fig.~\ref{method} provides an illustration of our workflow and Algorithm~\ref{alg:1} describes the design of the federated learning scheme. 
Algorithm~\ref{alg:1} provides a step-by-step explanation of our FL scheme.

% \begin{figure}[t!]
% \centering
% \includegraphics[width=0.7\columnwidth]{images/intro.png}
%     \caption{Overview of the proposed lightweight federated learning framework. For each client, there are two models, the student model and the frozen foundation model. The frozen foundation model generates the logit to guide the training of student models. Aggregated model by the server to aggregate information from different local models without observing their data. At the inference phase, the foundation model is discarded. }
%   \label{method}
% \end{figure}
\section{Experiments}
\label{sec:experiments}

\subsection{Experiments Setup}
\label{sec: expsetup}
\paragraph{Data Settings} 
We evaluate our algorithm on the CIFAR-10 dataset with 10 clients across seven data partitions at various levels of heterogeneity, including both IID and non-IID. 
For the non-IID data partitions we use (1) Dirichlet distribution, denoted as $\text{Dir}(\alpha)$ with $\alpha=1.0,\ 0.5,\ 0.1,\ 0.05,\ 0.01$; (2) Class Split, where each client's data is sampled from 2 of the 10 classes. 
We evaluate all algorithms over the balanced CIFAR-10 test set and report average accuracy over three trials to mitigate randomness of the runs.

\paragraph{Network Architectures}
For the choice of foundation models, we employ CLIP~\cite{radford2021learning} with backbones ViT-Base/32 (default) and RN50, while we use MobileNet-v2, EfficientNetB0, and ResNet18 as our proxy models.
In each of the proxy models, we replace the batch normalization layers with group normalization (8 groups) and train it from random initialization.

\paragraph{Training Setup and Hyper-parameters} 
% We evaluate our approach against FedAvg~\citep{mcmahan2017communication}, FedProx~\citep{li2020federated}, our implementation of centralized ProxyFL~\citep{kalra2023decentralized}/FML~\citep{DBLP:journals/corr/abs-2006-16765} with SGD. 
Throughout all algorithms and experiments, we use an SGD optimizer for training.
We train the proxy models for 600 epochs using a learning rate of 0.01, weight decay of 5e-4, and a step learning rate scheduler with a scale factor of 0.1 at epoch 200.
In ablation studies where we additionally consider directly fine-tuning foundation models, we train for 200 epochs with a learning rate of 2e-3, weight decay of 5e-4, and a cosine learning rate scheduler with 1 epoch of warmup. 
For comparisons against the SOTA algorithms we train the proxy models up to 500 epochs in FedAvg and FedProx, and 600 epochs in FML.

\begin{figure}[t!]
    
    \centering
    \begin{subfigure}[b]{0.3\columnwidth}
    \centering
    \includegraphics[width=\columnwidth]{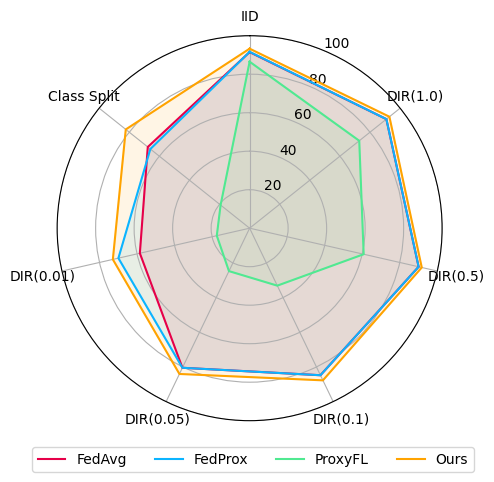}
    \caption{MobileNetV2}
    \label{fig:sota_mobilenetv2}
    \end{subfigure}
    \begin{subfigure}[b]{0.3\textwidth}
    \includegraphics[width=\columnwidth]{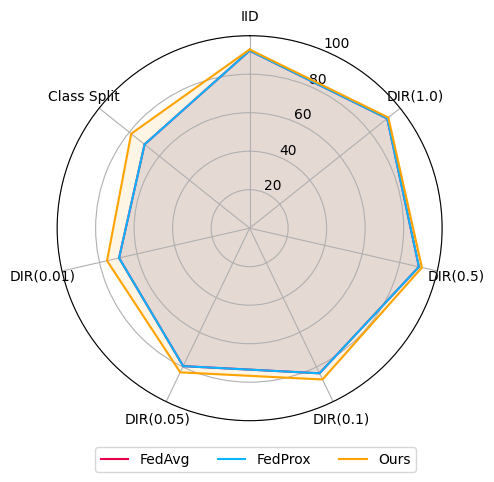}
    \caption{EfficientNetB0}
    \label{fig:sota_efficientnetb0}
    \end{subfigure}
    \begin{subfigure}[b]{0.3\textwidth}
    \includegraphics[width=\columnwidth]{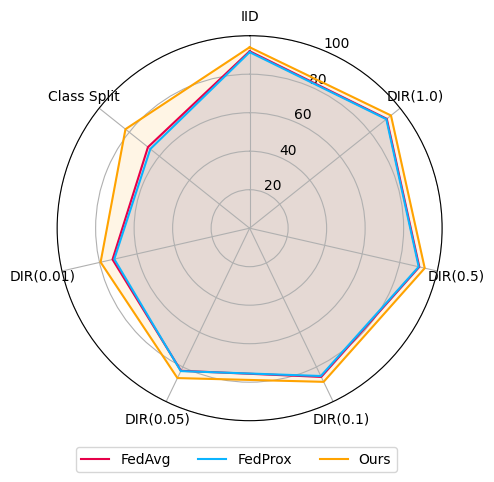}
    \caption{ResNet18}
    \label{fig:sota_resnet}
    \end{subfigure}
    \caption{Performance comparisons against existing works across a variety of data settings and proxy model backbones. \alg{} consistently outperforms prior art, by a large margin, under extremely heterogeneous data distributions. Larger area covered indicates a stronger FL approach.}
    \label{fig:sota}
\end{figure}

\subsection{Main results}
% \paragraph{Prior Art Comparison}
\paragraph{SOTA Algorithm Comparison}
We compare our approach against FedAvg~\citep{mcmahan2017communication}, FedProx~\citep{li2020federated}, and FML~\citep{DBLP:journals/corr/abs-2006-16765}, under multiple data heterogeneity partitions.
We visualize our results in Fig.~\ref{fig:sota}, where each data partition is represented as a vertex on the polar plot and accuracy is plotted along the radius.
From Fig.~\ref{fig:sota}, we observe that \alg{} robustly outperforms prior work across a variety of data distributions. \textit{In particular, our algorithm improves over FedProx (the best among prior art) by a wide margin, especially under the most extreme heterogeneous distributions (class split and dirichlet sampling with $\alpha=0.01, 0.05$).} 
% It presents the outstanding performance of Fed-LPFM to overcome client drift issues in FL. 
In addition, we highlight that using MobileNet as the backbone for both the private and proxy models, mimicing the setup in FML, performs poorly. 
We hypothesize that fine-tuning on the local data begins to bias the representations learned across both models, thus lending to significantly worsening performances as the data heterogeneity increases.

\paragraph{Proxy Model}
To establish the applicability of our approach to a variety of proxy model backbones, we evaluate across EfficienetNetB0, ResNet18, and MobileNetV2. 
We report and visualize the results in Figs.~\ref{fig:sota_efficientnetb0} and \ref{fig:sota_resnet}. 
% We clearly observe that our method consistently performs the best across the entire selection of proxy models.
% It shows consistent strength for our method cross different settings as observed before, in Fig. ~\ref{fig:sota_mobilenetv2}. 
We observe that our approach outperforms FedAvg and FedProx across the entire selection of proxy models under various data heterogeneity settings, especially the severe non-IID cases. 
% Moreover, the performances of our approach cross different data heterogeneity settings are robust under every proxy backbone settings tested.
In addition, we also observe that the improvement in performance from FedProx diminishes across both ResNet and EfficienNet, when compared to MobileNet. FedAvg and FedProx perform similar to one another.

% \begin{itemize}
%     \item Consistently, across all tested data distributions FedProx is the best performing method among existing works. 
%     \alg{} outperforms FedProx across all distributions, with a particularly wide margin under the most extreme heterogeneous distributions (e.g., class split, dirichlet sampling with $\alpha=0.01, 0.05$).
%     \item Interestingly, using MobileNet backbones for both private and proxy models, mimicing the setup in ProxyFL, performs the worst. We hypothesize that fine-tuning on the local data begins bias the representations learned across both models, thus lending to significantly worse performances as the heterogeneity of the data increases. 
%     \item \textcolor{red}{Comparison against ImageNet pre-trained model weights for MobileNet plus with and without BN/GN.}
% \end{itemize}

% \paragraph{Proxy Model Backbone}
% \begin{itemize}
%     \item We observe the improvement in performance offered by FedProx diminishes when applied to ResNet and EfficientNet. FedAvg and FedProx perform similar to one another.
%     \item Consistently, our approach boosts the performance of clients across all combinations of proxy model backbones and data heterogeneity settings.
% \end{itemize}

\begin{figure}[t!]
    \centering
    \begin{subfigure}{0.4\columnwidth}
    \includegraphics[width=0.97\columnwidth]{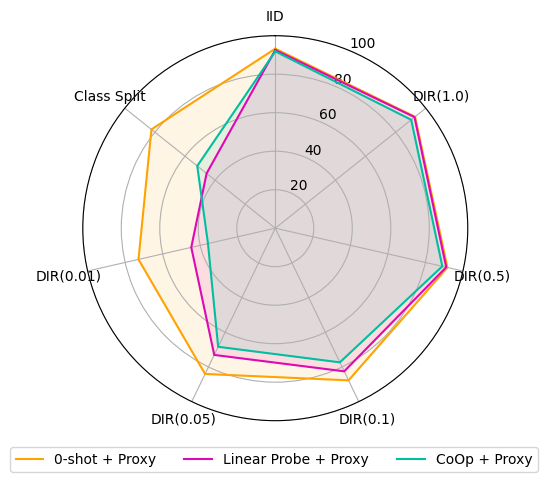}    
    \caption{Fine-tuning vs. 0-shot}
    \label{fig:ft_vs_0shot}
    \end{subfigure}
    \begin{subfigure}{0.4\columnwidth}
    \includegraphics[width=0.9\columnwidth]{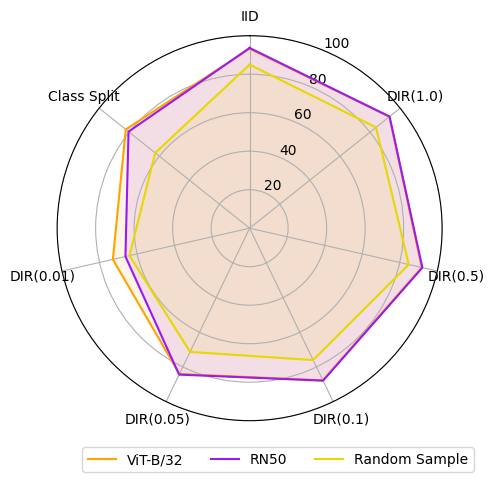}    
    \caption{Personalizing 0-shot Models}
    \label{fig:personalization}
    \end{subfigure}\\
    \begin{subfigure}{0.3\columnwidth}
    \includegraphics[width=0.92\columnwidth]{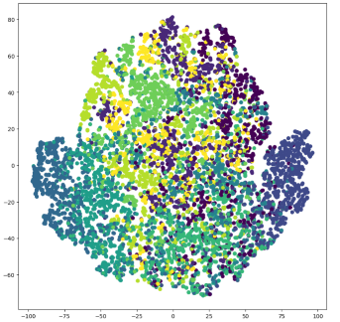}    
    \caption{Linear Probing}
    \label{fig:teacher_lp}
    \end{subfigure}
    \begin{subfigure}{0.3\columnwidth}
    \includegraphics[width=0.92\columnwidth]{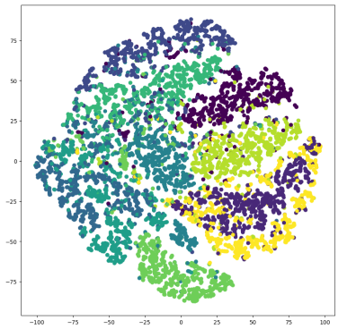}    
    \caption{Prompt Tuning}
    \label{fig:teacher_pt}
    \end{subfigure}
    \begin{subfigure}{0.3\columnwidth}
    \includegraphics[width=0.92\columnwidth]{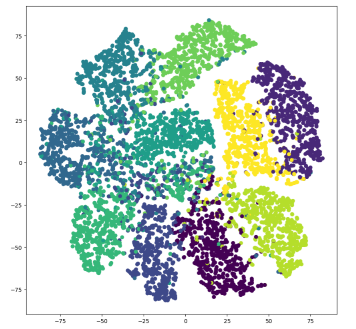}    
    \caption{0-shot}
    \label{fig:teacher_pt}
    \end{subfigure}
    \caption{(\textbf{\ref{fig:ft_vs_0shot}}) Fine-tuning foundation models on local data forces significantly worse performances under non-IID conditions. (\textbf{\ref{fig:personalization}}) Maintaining consistent foundation model backbones improves the synergy in information shared across clients. (\textbf{Bottom}) When compared to fine-tuned models, 0-shot models offer more diverse feature embeddings that reduce the bias of proxy models towards local data distributions.}
    \label{fig:ft_0shot_personalization}
\end{figure}

\paragraph{Fine-Tuned vs. 0-shot}
Our \alg{} method uses pre-trained foundation models with no available fine-tuning. 
To explore the impact of prior knowledge and how it affects distillation, we compare our 0-shot approach with first fine-tuning each client's foundation model(s) on local data (linear probing and prompt tuning).
Fig.~\ref{fig:ft_vs_0shot} illustrates how the 0-shot CLIP case outperforms the fine-tuned CLIP models. 
Our conjecture of this behavior is that \textit{fine-tuning the foundation model on local data results in a more personalized and biased knowledge representation which decreases the performance on a balanced test set.}
In addition, the more the data distribution is heterogeneous, the more knowledge encoded locally is personalized and biased. 
However, under more homogeneous settings there is a significant boost in the performance of clients when leveraging knowledge from both 0-shot and fine-tuned foundation models. 
We believe that the improvement shown when distilling from 0-shot models is largely due to the impact of strong diversity in its feature embeddings when compared to fine-tuned foundation models. 
We use tSNE plots to observe the spread of the encoded knowledge representations from foundation models.
From Fig.~\ref{fig:ft_0shot_personalization}, we can see that features from 0-shot foundation models cover a wider area when compared to fine-tuned models.

\paragraph{Personalizing Foundation Models}

% \begin{table}[h!]
% \centering
% \begin{tabular}{@{}cccccccc@{}}
% \toprule
% \multirow{2}{*}{} & \multirow{2}{*}{Class Split} & \multicolumn{5}{c}{Dirichlet} & \multirow{2}{*}{IID}\\ 
% & & \multicolumn{1}{c}{0.01} & \multicolumn{1}{c}{0.05} & 0.1 & 0.5 & 1.0 & \\
%                             \midrule
% CLIP: RN50       & 80.46 & 66.17 & 84.42 & 87.90 & 92.04 & 93.00 & 93.63\\ 
% CLIP: ViT-B/32   & 82.29 & 72.88 & 84.05 & 87.73 & 91.72 & 92.87 & 93.26 \\ 
% Random Selection & 62.75 & 64.02 & 71.33 & 76.05 & 84.82 & 83.99 & 85.01 \\ 
% \bottomrule
% \end{tabular}
% \end{table}

By keeping foundation model(s) private, \alg{} allows each client to personalize them.
From Fig.~\ref{fig:personalization}, we see that maintaining consistent backbones across the foundation models yields the highest improvement in performances while having a random sampling of backbones, between ViT-B/32 and RN50, forces a drop in performance.
We believe this behavior stems from following a naive strategy in combining the information presented by multiple proxy models.
The root of this behavior can be attributed to differences in knowledge/understanding of foundation models with disparate backbones. 
Instead, utilizing our approach along with personalized FL (\citep{t2020personalized, fallah2020personalized, li2021ditto, ghosh2020efficient, cho2021personalized}, etc.) could potentially boost the overall performance.

\section{Conclusions}
Overall, we establish \alg{} as an approach to leverage foundation models and help improve the performance and robustness achievable in small-scale models under the FL setting.
Distillation from pre-trained foundation models, as opposed to fine-tuned foundation models, provides the diversity in feature representations required to reduce the bias towards local distributions and thus, improve performance of clients across a variety of heterogeneous data distributions.
The use of logit-level distillation allows clients the flexibility to choose their local foundation models according to their individual constraints.
In doing so, we establish an important direction of future work; find an approach that, in a synergistic way, combines the information from disparate knowledge representations towards improved model performance.

% An important direction of future work is exploring a variety of metrics to assess a comprehensive way to identify noisy samples.
% In addition, we plan to expand on the contributions from multiple layers of the DNN to study the impact of feature hierarchy on the final performance. 
% Our goal is to jointly target PER in an effort to develop more cost and resource efficient training protocols, with a view to reducing the environmental impact of developing DNNs.

% What we have:
% Version A
% \begin{itemize}
%     \item As heterogeneity increases, CLIP:LP + MobileNet and CLIP:PT + MobileNet perform \textbf{worse} than FedAvg on MobileNet.
%     \item CLIP:0-shot + MobileNet \textbf{significantly improves} over all tested options across all settings.
% \end{itemize}

% What's next:
% \begin{itemize}
%     \item Version A: CLIP:0-shot + CLIP: LP ?? To showcase the observation is indeed valid across even FMFL?
%     \item Functioning Version B
%     \item Experiments on variable student backbones
%     \item More experimentation?
% \end{itemize}


\bibliographystyle{unsrtnat}
{
\bibfont
\bibliography{iclr2024_conference}
}

\newpage
\appendix
\section{Appendix}

The experimental results used to plot Figures in the main paper are shown in tabular form below. 
Our algorithm (i.e., \alg) consistently outperform other algorithms. 
We observed that \alg shows the largest improvement in the class split and Dirichlet distribution ($\alpha $ = 0.01 and 0.05), which are the most heterogeneous data distributions among different clients. 
For example, in the class split, \alg{} has a $21.6\%$ increase compared with Fedavg ($82.29\%$ vs $67.67\%$).

% \begin{table*}[h!] \label{tb1(a)}
% \centering
% \setlength{\tabcolsep}{3pt}
% \begin{tabular}{@{}cccccccc@{}}
% \toprule
% \multirow{2}{*}{Algorithms} & \multirow{2}{*}{Class Split} & \multicolumn{5}{c}{Dirichlet} & \multirow{2}{*}{IID}\\
% & & \multicolumn{1}{c}{0.01} & \multicolumn{1}{c}{0.05} & \multicolumn{1}{c}{0.1} & \multicolumn{1}{c}{0.5} & \multicolumn{1}{c}{1.0} & \\
%                             \midrule
% FedAvg & $67.67\pm 1.88$  & $58.49\pm15.72$ & $80.48\pm 1.07$ & $84.78\pm 1.65$ & $90.05\pm 0.21$ & $90.86 \pm 0.13$ & $91.48\pm 0.31$ \\ 
% FedProx     & $65.97\pm 5.07$ & $69.92\pm 2.07$ & $80.48\pm 1.07$ & $84.78\pm 1.65$ & $90.05\pm 0.21$ & $90.86\pm 0.13$ & $91.61\pm 0.34$ \\ 
% FML     & $19.29\pm 0.28 $ & $17.50\pm1.22$ & $24.75\pm 3.15$ & $33.13\pm2.58$ & $60.77\pm3.98$ & $72.84\pm1.56$ & $86.49\pm 0.17$\\ 
% \alg & $\bf 82.29\pm 1.12$ & {$\bf 72.88\pm 9.11$} & $\bf 84.05\pm 0.99$ & $\bf 87.73\pm 1.48$ & {$\bf 91.72\pm 0.31$} & {$\bf 92.87\pm 0.17$} & {$\bf 93.26\pm 0.17$} \\ 
% \bottomrule
% \end{tabular}
% \end{table*}

\begin{table}[h!]
\centering
\setlength{\tabcolsep}{3pt}
\begin{tabular}{@{}c|cccc@{}}
\toprule
Data Settings & FedAvg & FedProx & FML & \alg{} (MobileNetV2) \\
\midrule
Class Split & $67.67\pm 1.88$ & $65.97\pm 5.07$ & $19.29\pm 0.28$ & {$\bf 82.29\pm 1.12$} \\
Dir(0.01)   & $58.49\pm15.72$ & $69.92\pm 2.07$ & $17.50\pm 1.22$ & {$\bf 72.88\pm 9.11$} \\
Dir(0.05)   & $80.48\pm 1.07$ & $80.48\pm 1.07$ & $24.75\pm 3.15$ & {$\bf 84.05\pm 0.99$} \\
Dir(0.1)    & $84.78\pm 1.65$ & $84.78\pm 1.65$ & $33.13\pm 2.58$ & {$\bf 87.73\pm 1.48$} \\
Dir(0.5)    & $90.05\pm 0.21$ & $90.05\pm 0.21$ & $60.77\pm 3.98$ & {$\bf 91.72\pm 0.31$} \\
Dir(1.0)    & $90.86\pm 0.13$ & $90.86\pm 0.13$ & $72.84\pm 1.56$ & {$\bf 92.87\pm 0.17$} \\
IID         & $91.48\pm 0.31$ & $91.61\pm 0.34$ & $86.49\pm 0.17$ & {$\bf 93.26\pm 0.17$} \\
\bottomrule
\end{tabular}
\caption{The values of Figure 1(a) are shown in the current table.}
\label{tb1(a)}
\end{table}

% Figure 1(b)
% \begin{table*}[h!] \label{tb1(a)}
% \centering
% \setlength{\tabcolsep}{3pt}
% \begin{tabular}{@{}cccccccc@{}}
% \toprule
% \multirow{2}{*}{Algorithms} & \multirow{2}{*}{Class Split} & \multicolumn{5}{c}{Dirichlet} & \multirow{2}{*}{IID}\\
% & & \multicolumn{1}{c}{0.01} & \multicolumn{1}{c}{0.05} & \multicolumn{1}{c}{0.1} & \multicolumn{1}{c}{0.5} & \multicolumn{1}{c}{1.0} & \\
%                             \midrule
% FedAvg & $69.81\pm2.64$  & $69.56\pm 2.20$ & $79.53\pm0.95$ & $83.67\pm1.23 $ & $90.21\pm0.08 & $91.41\pm0.35 $ & $92.22\pm 0.17$ \\ 
% FedProx     & $69.81\pm 2.64$ & $69.56\pm 2.20$ & $79.53\pm0.95 $ & $83.67\pm1.23 $ & $90.21\pm0.08 $ & $91.41\pm0.35 $ & $92.22\pm 0.17$ \\ 
% % FML     & $\pm  $ & $\pm$ & $\pm $ & $\pm$ & $\pm$ & $\pm$ & $\pm $\\ 
% \alg \\(EfficientNetB0) & $\bf 78.70\pm 2.28$ & $\bf 75.98\pm1.29$ & $\bf 83.12\pm0.76 $ & $\bf 87.24\pm 1.46$ & $\bf 91.80\pm0.30 $ & $\bf 92.17\pm 0.09$ & $\bf 92.84\pm0.04 $\\ 
% \bottomrule
% \end{tabular}
% \end{table*}

Under different proxy model backbones, similar results can be observed. For both EfficientNet and ResNet case, we observed that Fed-LPFM outperforms other methods cross all data heterogeneity settings. 
\begin{table}[h!]
\centering
\setlength{\tabcolsep}{3pt}
\begin{tabular}{@{}c|ccc@{}}
\toprule
Data Settings & FedAvg & FedProx & \alg{} (EfficientNetB0) \\
\midrule
Class Split & $69.81\pm2.64$ & $69.81\pm2.64$ & {$\bf 78.70\pm2.28$} \\
Dir(0.01)   & $69.56\pm2.20$ & $69.56\pm2.20$ & {$\bf 75.98\pm1.29$} \\
Dir(0.05)   & $79.53\pm0.95$ & $79.53\pm0.95$ & {$\bf 83.12\pm0.76$} \\
Dir(0.1)    & $83.67\pm1.23$ & $83.67\pm1.23$ & {$\bf 87.24\pm1.46$} \\
Dir(0.5)    & $90.21\pm0.08$ & $90.21\pm0.08$ & {$\bf 91.80\pm0.30$} \\
Dir(1.0)    & $91.41\pm0.35$ & $91.41\pm0.35$ & {$\bf 92.17\pm0.09$} \\
IID         & $92.22\pm0.17$ & $92.22\pm0.17$ & {$\bf 92.84\pm0.04$} \\
\bottomrule
\end{tabular}
\caption{The values of Figure 1(b) are shown in the current table.}
\label{tb1(b)}
\end{table}

% The values of Figure 1(c) are shown in Table \ref{tb1(c)}. When we use ResNet18 as the lightweight model, our algorithms still get the best performance.  
% \begin{table*}[h!] \label{tb1(a)}
% \centering
% \setlength{\tabcolsep}{3pt}
% \begin{tabular}{@{}cccccccc@{}}
% \toprule
% \multirow{2}{*}{Algorithms} & \multirow{2}{*}{Class Split} & \multicolumn{5}{c}{Dirichlet} & \multirow{2}{*}{IID}\\
% & & \multicolumn{1}{c}{0.01} & \multicolumn{1}{c}{0.05} & \multicolumn{1}{c}{0.1} & \multicolumn{1}{c}{0.5} & \multicolumn{1}{c}{1.0} & \\
%                             \midrule
% FedAvg & $\pm$  & $\pm$ & $\pm$ & $\pm $ & $\pm & $\pm $ & $\pm $ \\ 
% FedProx     & $\pm $ & $\pm $ & $\pm $ & $\pm $ & $\pm $ & $\pm $ & $\pm $ \\ 
% FML     & $\pm  $ & $\pm$ & $\pm $ & $\pm$ & $\pm$ & $\pm$ & $\pm $\\ 
% \alg & $\bf \pm $ & {$\bf \pm $} & $\bf \pm $ & $\bf \pm $ & {$\bf \pm $} & {$\bf \pm$} & {$\bf\pm $} \\ 
% \bottomrule
% \end{tabular}
% \end{table*}

\begin{table}[h!]
\centering
\setlength{\tabcolsep}{3pt}
\begin{tabular}{@{}c|ccc@{}}
\toprule
Data Settings & FedAvg & FedProx & \alg{}(ResNet18) \\
\midrule
Class Split & $67.56 \pm 4.66$ & $65.99 \pm 6.41$ & {$\bf 82.50\pm 2.00$} \\
Dir(0.01)   & $73.06 \pm 2.10 $ & $72.13 \pm 1.38$ & {$\bf 79.41\pm 0.95$} \\
Dir(0.05)   & $82.22 \pm 0.79$ & $82.40\pm 0.46$ & {$\bf 86.42\pm 1.33$}\\
Dir(0.1)    & $85.74 \pm 1.38$ & $85.23\pm 0.89$ & ${\bf 88.59\pm 1.54}$ \\
Dir(0.5)    & $90.41\pm 0.30$ & $90.095\pm 0.54$ & {$\bf 93.30\pm 0.10$} \\
Dir(1.0)    & $91.13 \pm 0.13$ & $90.82\pm 0.25$ & {$\bf 93.92\pm 0.14$} \\
IID         & $91.94\pm 0.22$ & $91.45\pm 0.17$ & {$\bf 94.00\pm 0.20$}  \\
\bottomrule
\end{tabular}
\caption{The values of Figure 1(c) are shown in the current table.}
\label{tb1(c)}
\end{table}

% Figure 2(a)
% \begin{table}[h!]
% \centering
% \setlength{\tabcolsep}{3pt}
% \begin{tabular}{@{}cccccccc@{}}
% \toprule
% \multirow{2}{*}{Algorithms} & \multirow{2}{*}{Class Split} & \multicolumn{5}{c}{Dirichlet} & \multirow{2}{*}{IID}\\
% & & \multicolumn{1}{c}{0.01} & \multicolumn{1}{c}{0.05} & \multicolumn{1}{c}{0.1} & \multicolumn{1}{c}{0.5} & \multicolumn{1}{c}{1.0} & \\
%                             \midrule
% LP  & $45.63\pm 9.48$ & $44.76\pm 3.56$ & $73.07\pm 0.86$ & $82.54\pm 3.21$ & $91.08\pm 0.24$ & $92.53\pm 0.48$ & $92.56\pm 0.17$ \\ 
% PT   & $51.85\pm 4.58$ & $35.86\pm 6.03$ & $68.39\pm 1.65$ & $77.35\pm 3.75$ & $89.01\pm 0.12$ & $90.32\pm 0.33$ & $91.85\pm 0.23$ \\ 
% 0-shot & $\bf 82.29\pm 1.12$ & {$\bf 72.88\pm 9.11$} & $\bf 84.05\pm 0.99$ & $\bf 87.73\pm 1.48$ & {$\bf 91.72\pm 0.31$} & {$\bf 92.87\pm 0.17$} & {$\bf 93.26\pm 0.17$} \\ 
% \bottomrule
% \end{tabular}
% \end{table}
\begin{table}[ht!]
\centering
\setlength{\tabcolsep}{3pt}
\begin{tabular}{@{}c|ccc@{}}
\toprule
Data Settings & Linear Probing & Prompt Tuning & 0-shot (Ours) \\
\midrule
Class Split & $45.63\pm 9.48$ & $51.85\pm 4.58$ & {$\bf 82.29\pm 1.12$} \\
Dir(0.01)   & $44.76\pm 3.56$ & $35.86\pm 6.03$ & {$\bf 72.88\pm 9.11$} \\
Dir(0.05)   & $73.07\pm 0.86$ & $68.39\pm 1.65$ & {$\bf 84.05\pm 0.99$} \\
Dir(0.1)    & $82.54\pm 3.21$ & $77.35\pm 3.75$ & {$\bf 87.73\pm 1.48$} \\
Dir(0.5)    & $91.08\pm 0.24$ & $89.01\pm 0.12$ & {$\bf 91.72\pm 0.31$} \\
Dir(1.0)    & $92.53\pm 0.48$ & $90.32\pm 0.33$ & {$\bf 92.87\pm 0.17$} \\
IID         & $92.56\pm 0.17$ & $91.85\pm 0.23$ & {$\bf 93.26\pm 0.17$} \\
\bottomrule
\end{tabular}
\caption{The values of Figure 2(a) are shown in the current table.}
\label{tbl2(a)}
\end{table}

\begin{table}[h!]
\centering
\setlength{\tabcolsep}{3pt}
\begin{tabular}{@{}c|ccc@{}}
\toprule
Data Settings & CLIP: RN50 & CLIP: ViT-B/32 & Random Selection \\
\midrule
Class Split & $80.46\pm 4.08$ & {$\bf 82.29\pm 1.12$}  & $62.75\pm 3.24$ \\
Dir(0.01)   & $66.17\pm 1.19$ & {$\bf 72.88\pm 9.11$}  & $64.02\pm 3.52$ \\
Dir(0.05)   & {$\bf 84.42\pm 0.41$} & $84.05\pm 0.99$  & $71.33\pm 1.52$ \\
Dir(0.1)    & {$\bf 87.90\pm 1.58$} & $87.73\pm 1.48$  & $76.05\pm 3.53$ \\
Dir(0.5)    & {$\bf 92.04\pm 0.25$} & $ 91.72\pm 0.31$ & $84.82\pm 0.45$ \\
Dir(1.0)    & {$\bf 93.00\pm 0.10$} & $ 92.87\pm 0.17$ & $83.99\pm 0.99$ \\
IID         & {$\bf 93.63\pm 0.30$} & $93.26\pm 0.17$  & $85.01\pm 0.47$ \\
\bottomrule
\end{tabular}
\caption{The values of Figure 2(b) are shown in the current table.}
\label{tbl2(b)}
\end{table}
As supporting materials for Fig.~\ref{fig:ft_vs_0shot}, we report the numerical results for testing fine-tuned CLIP (linear probing and prompt tuning) and zero-shot CLIP cross different data heterogeneity levels. We observed that zero-shot CLIP offers best and more robust performances when compared to fine-tuned methods.

% \begin{table}[h!]
% \centering
% \begin{tabular}{@{}cccccccc@{}}
% \toprule
% \multirow{2}{*}{} & \multirow{2}{*}{Class Split} & \multicolumn{5}{c}{Dirichlet} & \multirow{2}{*}{IID}\\ 
% & & \multicolumn{1}{c}{0.01} & \multicolumn{1}{c}{0.05} & 0.1 & 0.5 & 1.0 & \\
%                             \midrule
% CLIP: RN50       & $80.46\pm 4.08$ & $66.17\pm 1.19$ & $\bf 84.42\pm 0.41$ & $\bf 87.90\pm1.58$ & $\bf 92.04\pm0.25$ & $\bf93.00\pm 0.10$ & $\bf 93.63\pm0.30$\\ 
% CLIP: ViT-B/32   & $\bf 82.29\pm 1.12 $ & $\bf 72.88 \pm 9.11$ & $84.05\pm 0.99$ & $87.73\pm 1.48$ & $91.72\pm0.31$ & $92.87\pm0.17$ & $93.26\pm 0.13$ \\ 
% Random Selection & $62.75\pm 3.24$ & $64.02\pm 3.52$ & $71.33\pm 1.52$ & $76.05\pm 3.53$ & $84.82\pm 0.45$ & $83.99\pm 0.99$ & $85.01\pm 0.47$ \\ 
% \bottomrule
% \end{tabular}
% \end{table}

As supporting materials for Fig. ~\ref{fig:personalization}, we report the results of using CLIP: ResNet50 and CLIP:ViT-B/32 as well as random sampling of them, with uniform prior, as foundation models. It shows that random selection of pre-trained models offers worst performances when compared to the other two.

\end{document}